\documentclass{WileyMSP-template}

\usepackage[utf8]{inputenc}  
\usepackage[T1]{fontenc}     

\usepackage{moreverb,url}
\usepackage{graphicx}
\usepackage{svg}
\usepackage{subfig}
\usepackage{float}
\usepackage{array}
\usepackage{makecell}
\usepackage{pifont}
\usepackage{enumitem}
\usepackage[export]{adjustbox}
\usepackage[flushleft]{threeparttable}
\usepackage{booktabs}
\usepackage{microtype}
\usepackage{ragged2e}
\justifying

\usepackage{amsmath,amssymb,amsbsy,amsthm}
\usepackage{mathtools}
\mathtoolsset{showonlyrefs} 
\usepackage{mathrsfs}

\theoremstyle{definition}

\usepackage[numbers]{natbib}
\let\citet\cite
\let\citep\cite

\usepackage{titlesec}
\renewcommand{\theparagraph}{\alph{paragraph})}
\titleformat{\paragraph}[hang]{\normalfont\normalsize\itshape}{\hspace{1em}\theparagraph}{0.5em}{}
\titlespacing*{\paragraph}{0pt}{\baselineskip}{0em}

\let\oldsubsubsection\subsubsection
\renewcommand{\subsubsection}[1]{\oldsubsubsection{#1}\mbox{}\par}

\usepackage{algorithm}
\usepackage{algorithmicx}
\usepackage{algpseudocode}
\usepackage{setspace}

\PassOptionsToPackage{hyphens}{url} 
\usepackage[colorlinks=true,
            linkcolor=blue,
            filecolor=magenta,
            citecolor=magenta,
            urlcolor=magenta,
            bookmarksopen,
            bookmarksnumbered]{hyperref}

\DeclareMathOperator*{\argmin}{arg\,min}

\let\oldmaketitle\maketitle
\renewcommand{\maketitle}{%
  \oldmaketitle
  \setcounter{footnote}{2}%
}

\begin{document}

\pagestyle{fancy}
\rhead{\vspace{0.3cm}}

\twocolumn[
  \begin{@twocolumnfalse}
    \title{Scalable Task Planning via Large Language Models and Structured World Representations}
    \maketitle

    \author{Rodrigo P\'{e}rez-Dattari$^1$,}
    \author{Zhaoting Li$^{1}$,}
    \author{Robert Babu\v{s}ka$^{1,2}$}
    \author{Jens Kober$^1$}
    \author{and Cosimo Della Santina$^{1,3}$,}
    
    \begin{affiliations}
        $^1$Department of Cognitive Robotics, Delft University of Technology, Netherlands.\\
        $^2$Czech Institute of Informatics, Robotics, and Cybernetics, Czech Technical University in Prague, Czech Republic.\\
        $^{3}$Institute of Robotics and Mechatronics, German Aerospace Center (DLR), Wessling, Germany.\\
    \end{affiliations}
    
  \end{@twocolumnfalse}
]

\begin{abstract}
\textbf{Abstract---}
Planning methods struggle with computational intractability in solving task-level problems in large-scale environments.
This work explores leveraging the commonsense knowledge encoded in LLMs to empower planning techniques to deal with these complex scenarios. We achieve this by efficiently using LLMs to prune irrelevant components from the planning problem's state space, substantially simplifying its complexity.
We demonstrate the efficacy of this system through extensive experiments within a household simulation environment, alongside real-world validation using a 7-DoF manipulator (video \url{https://youtu.be/6ro2UOtOQS4}).
\end{abstract}

\keywords{Task Planning, LLMs, Graphs, Taxonomy, Robotics}

\section{Introduction}
\label{sec:introduction}
In homes, hospitals, and factories, there is an increasing demand for robotic assistants that can react to high-level commands like ``\textit{Clean room A.}" 
Search-based planning \cite{bonet2001planning,hoffmann2001ff,helmert2006fast} provides a promising framework for generating the action sequences necessary to solve such tasks. However, these techniques quickly reach their computational limits when applied to the realistic scenarios mentioned above, as these scenarios are populated by a large number of objects with which the robot can interact in multiple ways.

Pre-trained Large Language Models (LLMs) are gaining popularity as a zero-shot alternative for generating plans from language instructions. However, the generated plans can often be unrealistic, unfeasible, or even wrongly formulated. This issue can be mitigated, yet not solved, by providing the LLM with detailed environmental information in its context window, including task constraints and robot capabilities~\cite{rana2023sayplan,vemprala2024chatgpt,zhao2024large,wang2024survey}.
Similarly to the classic search approaches, these methods do not scale well to large settings, as considerably extending the context window can potentially lead to inaccuracies \cite{liu2024lost} and higher operational costs.
\begin{figure}[t]
    \centering
    \includegraphics[trim = {0 20 0 0}, clip, width=\columnwidth]{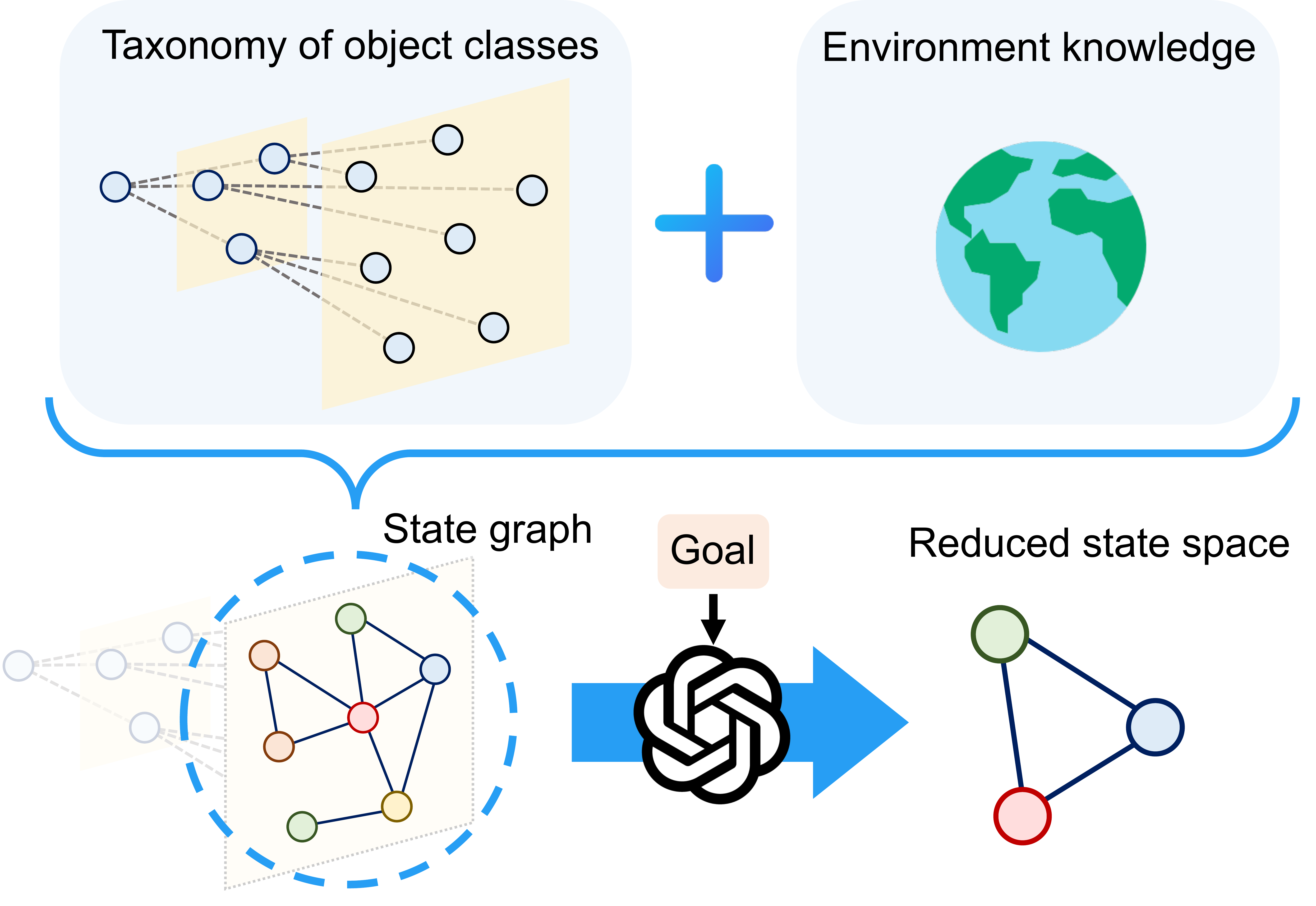}
    \caption{Summary of the proposed framework. A taxonomy of object classes, where the lower level represents objects (e.g., a computer) and the higher levels represent groups of objects (e.g., electronics), is combined with knowledge of the environment to create a state graph that indicates attributes of objects (e.g., A is open) and relationships between them (e.g., A is inside B). For a given task goal, an LLM uses the object taxonomy, the state graph, and its commonsense knowledge to derive a reduced graph that contains only the necessary objects to achieve the task planning problem.}
    \label{fig:cover}
\end{figure}
Here, we propose to use LLMs to address the scalability challenge by reducing the size of the problem before planning begins. This results in a scalable and effective task-planning framework. We achieve this by endowing LLMs with a structured world representation.
The key idea behind this work is that --- despite the presence of thousands of objects in complex environments --- only a few are essential for specified tasks.
Data-driven methods have already been considered in this context, with models trained to predict object relevance \cite{silver2021planning}.
However, past efforts have been hindered by the need for substantial retraining when presented with new environments.

Instead, we leverage the inherent commonsense knowledge of LLMs to identify relevant objects based on a specified task objective. To ground these predictions in the real world, we integrate LLMs with a $\text{graph-based}$ environment representation, where nodes represent objects and their attributes, and edges denote relationships between objects. For example, a graph could indicate that the node \emph{oven} has attribute \emph{is closed} and that it has an edge with the node \emph{cake}, indicating that the cake \emph{is inside} the oven. Such object-centric representations facilitate encoding prior knowledge and structural information about the world using language, being directly accessible to LLMs. 

Additionally, removing objects from the world representation is simple, as it only requires removing nodes from the graph. Lastly, to efficiently feed this information to the LLM without requiring extensive context windows, we propose to organize objects hierarchically using a taxonomy, resulting in a condensed graph with which the LLM can interact. Fig.~\ref{fig:cover} summarizes the introduced concepts.

To validate our method, experiments were conducted using the \emph{VirtualHome} simulator \cite{puig2018virtualhome}, which contains environments with hundreds of objects, and a real-world 7-DoF robotic manipulator tasked with rearranging a configuration of objects on a table. This paper is structured as follows: Section~\ref{sec:related_works} discusses related works in task planning and problem size reduction; Section~\ref{sec:preliminaries} outlines the preliminary concepts and definitions; Section~\ref{sec:problem_formulation} formulates the problem our method addresses;
Section~\ref{sec:methodology} details our proposed methodology; Section~\ref{sec:experiments} presents the experimental setup and results; and finally, Section~\ref{sec:conclusions} concludes the paper and discusses future work directions.

\section{Related Work}\label{sec:related_works}
%
The problem of Task Planning (TP) has been extensively studied. Traditionally, planning problems are represented using languages like STRIPS \cite{fikes1971strips} or PDDL \cite{ghallab1998} and solved with tree search methods \cite{bonet2001planning,hoffmann2001ff,helmert2006fast}. In robotics, these methods have evolved to handle continuous variables and geometric information, leading to Task and Motion Planning (TAMP) \cite{garrett2020pddlstream,15-toussaint-IJCAI,garrett2021integrated}. However, as planning problems grow, TP and TAMP approaches become intractable \cite{garrett2021integrated, agia2022taskography, silver2021planning}. While methods like Monte Carlo Tree Search (MCTS) \cite{coulom2006efficient} scale better, they still struggle with large-scale problems.
%
%
To make large planning problems manageable, \cite{silver2021planning} proposes learning the importance of objects relative to a task goal and then solving the problem using only a subset of objects. Similarly, \cite{gnad2019learning} identifies relevant actions to reduce the search space. However, these methods require prior examples from the environment to learn relevance, limiting zero-shot applicability. To address this, \cite{zhao2024large} proposes leveraging the commonsense knowledge of LLMs to bias planning algorithms, where an LLM is queried to select the most relevant actions for a given state and task goal.

LLMs have gained popularity for solving planning problems due to their ability to embed commonsense knowledge and infer reasonable plans for tasks \cite{wang2024survey}. However, challenges remain: without specific task information, these models may hallucinate infeasible solutions for robots. To address these challenges, various works propose grounding robot plans in specific domains using methods like affordance functions \cite{brohan2023can,huang2022language}, semantic distance minimization \cite{huang2022language,zhao2024large}, and feasibility checks from world representations \cite{rana2023sayplan,singh2023progprompt}. These approaches assume a library of motion primitives \cite{latombe2012robot,saveriano2023dynamic,billard2022learning} for querying high-level planning solutions, ranging from learning-based methods \cite{open_x_embodiment_rt_x_2023,figueroa2018physically,perez2023stable} to operational space control and motion planning \cite{siciliano2009robotics,spong2020robot}. This separation of TP from motion planning is limited when geometrical aspects are critical, as current LLMs show limited trajectory-level operation \cite{chen2023autotamp}. Some works suggest translating natural language tasks into formal planning languages for complete TAMP problem-solving \cite{chen2023autotamp,liu2023llm+}, or generating multiple candidate plans and selecting the one with the best geometrical feasibility \cite{lin2023text2motion}. However, despite the importance of addressing the complete TAMP problem with LLMs, our work focuses on TP leveraging a library of motion primitives. Expressive motion primitives can address challenging problems \cite{brohan2023can,singh2023progprompt,huang2022inner}, and future extensions to TAMP problems are possible.

%
The closest works to ours involve combining graphs with spatial structure to build collapsed state representations, which are then locally expanded to find a plan \cite{rana2023sayplan,rajvanshi2024saynav}. Such works continue the trend of approaches aimed at reducing problem size using 3D scene graphs \cite{armeni20193d,Rosinol20rss-dsg}, which were extended in \cite{agia2022taskography} for usability in planning problems. Lastly, some of these ideas have been expanded into the TAMP domain as well \cite{ray2024task}. Unlike these methods, our approach centers on large-scale planning, and it is not specifically tailored for 3D scene graphs. Moreover, our contribution focuses on reducing the size of the state space rather than planning.
\section{Preliminaries}
\label{sec:preliminaries}
\subsection{Task Planning}
Consider a setting with discrete states $s_t \in \mathcal{S}$ and actions $a_t \in \mathcal{A}$. Furthermore, these variables are described in a discrete-time framework, where the subscript $t$ denotes the corresponding time step. For any given state, a known set of \emph{applicable actions} ${\bar{\mathcal{A}}(s_{t})\subseteq \mathcal{A}}$ exists, such that ${\bar{\mathcal{A}}:\mathcal{S} \twoheadrightarrow\mathcal{A}}$, where the symbol $\twoheadrightarrow$ defines a set-valued function, i.e., elements of $\mathcal{S}$ are mapped to subsets of $\mathcal{A}$. This set also considers the robot's capabilities; for example, a single-armed robot holding a banana cannot grab an apple since its hand is already occupied. However, a dual-armed robot could simultaneously grab the apple. This concept is also known as \emph{affordance} \cite{brohan2023can,khetarpal2020can}. Then, given a state and a selected action from $\bar{\mathcal{A}}$, a known \emph{transition function} $f:\mathcal{S} \times \mathcal{A} \to \mathcal{S}$, dictates the evolution of our environment/system from a time step to the next one, i.e., $f(s_{t}, a_{t})=s_{t+1}$.

In this setting, a planning problem includes an initial state $s_{0} \in \mathcal{S}$ and a set of goal states $\mathcal{G} \subseteq \mathcal{S}$ \cite{garrett2021integrated}. The set $\mathcal{G}$ represents a collection of states that satisfy a specific desired condition. For instance, a desired condition in a household might be having clean dishes. However, multiple household configurations can result in clean dishes. Therefore, all such configurations would constitute $\mathcal{G}$. As a result, the objective is to obtain a policy $\pi:\mathcal{S} \to \mathcal{A}$, that from any given $s_{0}$, by continuously interacting with the environment, generates a trajectory $\tau = (s_{0}, ..., s_{n})$, where $s_{t+1} = f(s_{t}, \pi(s_{t}))$, $\pi(s_{t}) \in \bar{\mathcal{A}}(s_{t})$, and $n \in \mathbb{N}_{0}$, such that $s_{n} \in \mathcal{G}$.

\subsection{Graph-based State Representation}
As the name suggests, a graph-based state representation represents the state of the environment via a graph~\cite{puig2018virtualhome,agia2022taskography,zhao2024large}. Such graphs are tuples $S=(O,R)$ that model relationships $R$ between objects $O$. Objects, represented as the nodes or vertices of the graph, are the entities in our environment, such as the kitchen table, a robot, or the bedroom. The relationships between the objects, represented as the edges of the graph, correspond to physical relationships, such as \emph{robot in the bedroom} or \emph{banana is being grabbed by the robot}. Given a set of classes $\mathcal{C}$ and attributes $\mathcal{H}$, each object has the form $o_{i}=(c_{i}, h_{i})$, where $c_{i} \in \mathcal{C}$ is the class of the object, e.g., a microwave, and $h_{i} \in \mathcal{H}$ is a vector of attributes that provide information about the intrinsic state of the object, e.g., \emph{on/off} or \emph{open/closed}. As a result, we have the set $O = \{o_{1}, \ldots, o_{m}\}$, with $m \in \mathbb{N}$. Moreover, given a set of possible relationships between objects $\mathcal{R}$, every existing one is defined as $r_{i}=(o_{j}, o_{k}, k_{i})$, where $k_{i} \in \mathcal{R}$ represents relationships such us \emph{grabbed} or \emph{inside}. Consequently, the set of relationships/edges of the graph corresponds to ${R = \{r_{1}, \ldots, r_{l}\}}$, with $l \in \mathbb{N}$. 

This construction allows representing the environment's state $s_{t}$ by directly employing the graph $S$, where $O$ contains the information about each object and its attributes, and $R$ represents every existing relationship between objects.

\section{Problem Formulation}
\label{sec:problem_formulation}
Given an initial environment graph $S_{0}$ and a set of goal states $\mathcal{G}$, our objective is to reduce the dimensionality of the state-action space, rendering the problem solvable. Specifically, we aim to derive a subgraph $\bar{S}_{0} = (\bar{O}_{0}, \bar{R}_{0}) \subseteq S_{0}$, minimizing the number of objects $|O|$ in the environment, while \emph{maintaining all necessary objects} $O_{\mathrm{g}}$ to achieve the task goal. Importantly, the set $O_{\mathrm{g}}$ depends not only on $\mathcal{G}$ but also on the initial object relationships $R_{0}$. For example, consider the task of placing an apple on a table, with the apple initially inside a closed fridge. Although the fridge's state is irrelevant to $\mathcal{G}$, it is imperative in $O_{\mathrm{g}}$, as the task requires interacting with it. Thus, $O_{\mathrm{g}}$ is influenced by the initial graph's objects and relationships. 

We define the problem as follows:
\begin{equation}
\label{eq:problem}
\bar{O}_{0} = \argmin_{O \subseteq O_{0}} |O|, \quad
\text { s.t. } O_{\mathrm{g}} \subseteq O, \, \footnotemark
\end{equation} 
where the subscript in $\bar{O}_{0}$ can be dropped, i.e., $\bar{O}$, as the number of objects is invariant over time.
Notably, addressing this problem significantly reduces the number of edges at any time step $|\bar{R}_{t}|$, as only relationships involving objects in $\bar{O}$ are relevant.
\footnotetext{While binary variable formulations are generally preferred to the $\subseteq$ notation, we use $\subseteq$ for its concise representation of the problem.}
\section{Method}\label{sec:methodology}
\begin{figure}[t]
    \centering
    \includegraphics[trim = {0 20 0 0}, clip, width=\columnwidth]{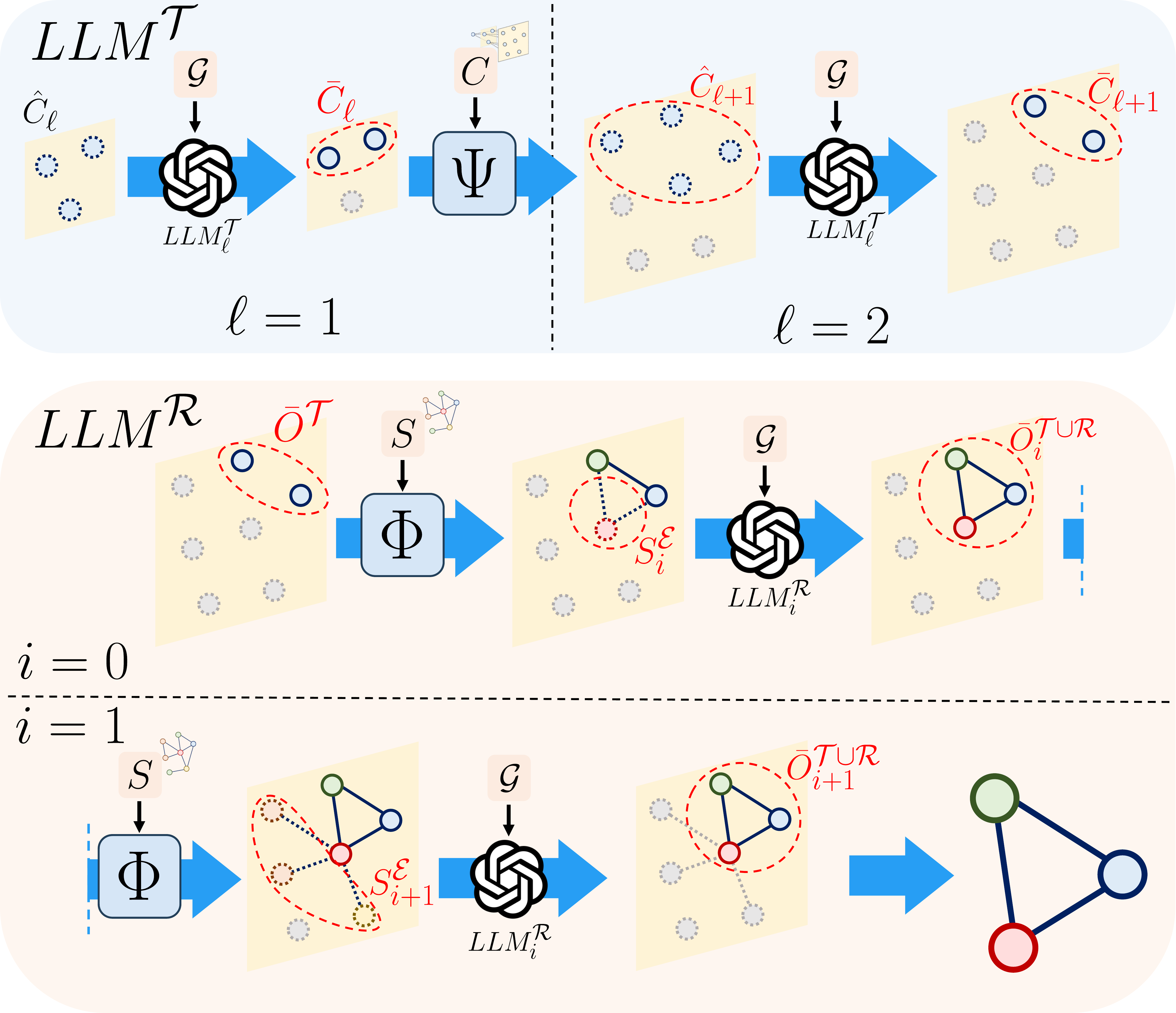}
    \caption{Example of the proposed method. \textbf{Step 1 ($LLM^{\mathcal{T}}$):} Relevant objects are selected from an object taxonomy $C$. At the highest hierarchical level ($\ell=1$), three object categories are provided to the LLM, which selects two as relevant ($\bar{C}_{\ell}$). The child nodes of $\bar{C}_{\ell}$ ($\hat{C}_{\ell+1}$), obtained via $\Psi$, are then provided to the LLM to further select the relevant ones ($\bar{C}_{\ell+1}$).\newline
    \textbf{Step 2 ($LLM^{\mathcal{R}}$):} Relevant relationships are iteratively selected using the graph state $S$. In the first iteration ($i=0$), the objects obtained from Step 1 ($\bar{O}^{\mathcal{T}}$) are fed into the function $\Phi$, which locally expands the graph to identify objects interacting with $\bar{O}^{\mathcal{T}}$. One object is identified ($S^{\mathcal{E}}_{i}$). Subsequently, the LLM determines that the interacting object is relevant. A second iteration starts ($i=1$), where the new graph $\bar{O}_{i}^{\mathcal{T}\cup\mathcal{R}}$ is expanded with interacting objects $S^{\mathcal{E}}_{i+1}$. However, none of these objects are deemed relevant by the LLM, concluding the object selection process.}
    \label{fig:method}
\end{figure}

We combine two ingredients to solve the problem outlined in \eqref{eq:problem}: knowledge of the environment's structure (as represented in graph states), and commonsense knowledge (as provided by pre-trained LLMs). To achieve this, we follow two steps, exemplified in Fig. \ref{fig:method}.

\subsection{Step 1: Environment-agnostic Object Selection}
First, we propose to obtain a subset of objects $\bar{O}^{\mathcal{T}}$ only as a function of the initial set of objects present in the environment $O$ and the task objective $\mathcal{G}$, disregarding any attribute or relationships the objects have. To achieve this, we query an LLM to select from $O$ the objects that are potentially relevant for solving the task. 
As a result, the LLM can be represented as the function $LLM^{\mathcal{T}}:(O, \mathcal{G}) \mapsto \bar{O}^{\mathcal{T}}$. 

Unfortunately, as discussed previously, in large-scale problems, it is impractical, expensive or unfeasible to directly feed the complete set of objects to the LLM. To address this, we incorporate a taxonomy of object classes that provides a collapsed representation of $O$ with which the LLM can interact, efficiently exploring it to gain access to the most promising groups of objects present in $O$.

\subsubsection{Taxonomy Graph States}
The taxonomy can be represented as a graph and, therefore, can be seen as an extension to the previously introduced graph state $S$. Generally, taxonomies categorize items hierarchically into groups or types. Incorporating a taxonomy into our graph allows us to collapse multiple classes into broader concepts. Hence, we extend the graph $S$ with a collection of \emph{taxonomy nodes} $C$, which group together object classes and/or categories represented in other taxonomy nodes lower in the hierarchy. For instance, an object of the class \emph{keyboard} can be grouped into the category \emph{computing}, represented by a taxonomy node. Simultaneously, the category \emph{computing} can be grouped into a larger category, \emph{electronics}, represented by another taxonomy node. Then, to categorize any object or taxonomy node, we incorporate edges $T$ connecting child nodes (objects or taxonomy nodes) with parent nodes (taxonomy nodes) representing their corresponding categories. Consequently, we get the extended graph $E = (O \cup C, R \cup T)$, where $S \subseteq E$, which we refer to as a \emph{taxonomy graph state}.
%

\subsubsection{Interacting with the taxonomy}
For the LLM to interact with the taxonomy, it first selects the categories, i.e., taxonomy nodes, that are most relevant to the task. Since the categories are hierarchical, the LLM initially only receives the higher-level categories and identifies the most pertinent ones according to the task objective $\mathcal{G}$. Then, by identifying the edges of these selected categories, the algorithm descends one level in the hierarchy. The information is once again fed into the LLM to select the most relevant lower-level nodes. This process repeats until further descent in the hierarchy is not possible, leaving only the object nodes $\bar{O}_{\mathcal{T}}$.

More formally, at any given hierarchical level $\ell$ from $C$ we have the set of nodes $C_{\ell} \subseteq C$. Here,
$\ell=1$ is the highest hierarchical level and increasing $\ell$ represents lower hierarchical levels. Assuming nodes have already been selected at an upper level $\ell-1$, denoted by ${\bar{C}_{\ell-1} \subseteq C_{\ell-1}}$, from the nodes at $\ell$ that are connected to $\bar{C}_{\ell-1}$, called $\hat{C}_{\ell}$, the LLM must select the relevant nodes $\bar{C}_{\ell}$. This is represented by the function
\begin{equation}
LLM_{\ell}^{\mathcal{T}}: (\hat{C}_{\ell}, \mathcal{G}) \mapsto \bar{C}_{\ell},
\end{equation}
which is iterated until reaching the lowest hierarchy level $\ell=L$. This iterative process requires a function $\Psi$ that, given the taxonomy $C$ and nodes selected at $\ell$, $\bar{C}_{\ell}$, outputs the nodes connected to $\bar{C}_{\ell}$ lower in the hierarchy $\hat{C}_{\ell+1}$, i.e.,
\begin{equation}
\Psi: (C, \bar{C}_{\ell}) \mapsto \hat{C}_{\ell+1}.
\end{equation}
The usage of $\Psi$, along with the complete process of computing $LLM^{\mathcal{T}}$, is detailed in Algorithm \ref{algorithm:LLM^T}.
\begin{algorithm}[h]
\caption{$LLM^{\mathcal{T}}$: Environment-agnostic selection}\label{algorithm:LLM^T}
\begin{algorithmic}[1]
\State \textbf{Require}: taxonomy $C$, goal $\mathcal{G}$, and functions $\Psi$, $LLM^{\mathcal{T}}_{\ell}$
\State \textbf{initialize} $\bar{C}_{\ell} \leftarrow C_{0}$ \quad \quad \quad \quad \quad \quad \quad\# Initializing node
\For{$\ell \in \{1, 2, \ldots, L-1\}$}
    \State \textbf{get lower nodes} $\hat{C}_{\ell+1} \leftarrow \Psi(C, \bar{C}_{\ell})$
    \State \textbf{select nodes} $\bar{C}_{\ell+1} \leftarrow LLM_{\ell}^{\mathcal{T}}(\hat{C}_{\ell+1}, \mathcal{G})$
    \State \textbf{assign} $\bar{C}_{\ell} \leftarrow \bar{C}_{\ell+1}$
\EndFor
\State \textbf{output} $\bar{O}^{\mathcal{T}} \leftarrow \bar{C}_{L}$
\end{algorithmic}
\end{algorithm}

We assume that a taxonomy of the objects exists, aligning with the setting of this work: despite the large scale of the problem, there is knowledge about the environment in which the robot is operating. Consequently, it is possible to construct a taxonomy. This process could also be aided by an LLM, though this aspect falls outside the scope of this work. Furthermore, in practice, the function $\Psi$ can be easily implemented.

\subsection{Step 2: Relationship-based Object Selection}
Up to this point, the introduced method selects a subset of relevant objects without considering their interactions within the environment (Step 1). Next, we describe how to incorporate interaction information into the selection process (Step 2), which can often reveal additional relevant objects to consider. 

Step 2 grounds the LLM predictions of Step 1 on object interactions that are specific to the environment at hand. For example, consider a scenario where the objective is to place a banana on the kitchen table. From this objective, we expect the LLM to select the objects \emph{banana} and \emph{kitchen table} in Step 1. However, by examining the graph neighborhood of the kitchen table node, it can be observed that every banana has an edge to a \emph{plastic container}, as they are stored \emph{inside} it. Therefore, the container should also be selected as a relevant object, as it must be incorporated into the planning problem to fulfill the task.

As a result, in Step 2, based on the relationships present in the initial graph state $S_{0}$, the neighborhoods of the nodes selected in Step 1, i.e.,  $\bar{O}^{\mathcal{T}}$, are explored. We introduce the function $\Phi$ that, given $S_{0}$ and $\bar{O}^{\mathcal{T}}$, outputs the \emph{connected subgraph} $S^{\mathcal{E}}=(O^{\mathcal{E}}, R^{\mathcal{E}})$. The graph $S^{\mathcal{E}}$ is composed of the objects in $O$ sharing edges with $\bar{O}^{\mathcal{T}}$, referred to as the \emph{connected objects} $O^{\mathcal{E}}$, and the shared edges, $R^{\mathcal{E}}$. Hence,
\begin{equation}
    \Phi: (S_{0}, \bar{O}^{\mathcal{T}}) \mapsto S^{\mathcal{E}}. 
\end{equation}
The subgraph $S^{\mathcal{E}}$, the objects selected in Step 1, $\bar{O}^{\mathcal{T}}$, and the task objective $\mathcal{G}$, are fed to an LLM that is queried to select from the connected objects those relevant for achieving the objective based on their relationships $R^{\mathcal{E}}$ with $\bar{O}^{\mathcal{T}}$. We represent this process with the function
\begin{equation}
    LLM^{\mathcal{R}}:(\bar{O}^{\mathcal{T}}, S^{\mathcal{E}}, \mathcal{G}) \mapsto \bar{O}^{\mathcal{R}},
\end{equation}
where $\bar{O}^{\mathcal{R}}$ is new set of objects selected as relevant.

\subsubsection{Iterative selection process}
Importantly, the introduced selection process should be repeated through multiple iterations, as the newly selected objects might also have relationships that need to be accounted for. For example, the plastic container might be related to the \emph{fridge} because it is stored inside it. Thus, the fridge should also be incorporated into our problem. To determine the times $LLM^{\mathcal{R}}$ must be employed, we can iterate until the LLM judges that none of the newly connected objects are relevant for solving the problem or when a maximum number of iterations $I$ is reached. 

Following our previous notation, and by incorporating subindices $i$, which denote the relationship-based selection iteration, the previously defined LLM function can be modified to
\begin{equation}
    {LLM_{i}^{\mathcal{R}}:(\bar{O}_{i}^{\scriptscriptstyle \mathcal{T} \cup \mathcal{R}}, S_{i}^{\mathcal{E}}, \mathcal{G}) \mapsto \bar{O}_{i}^{\mathcal{R}}}.
\end{equation}
Here, ${\bar{O}_{i}^{\scriptscriptstyle \mathcal{T} \cup \mathcal{R}} = \bar{O}_{i-1}^{\scriptscriptstyle \mathcal{T} \cup \mathcal{R}} \cup \bar{O}_{i-1}^{\mathcal{R}}}$, where $\bar{O}_{0}^{\scriptscriptstyle \mathcal{T} \cup \mathcal{R}}=\bar{O}^{\mathcal{T}}$, and $S_{i}^{\mathcal{E}} = \Phi(S_{0}, \bar{O}_{i}^{\scriptscriptstyle \mathcal{T} \cup \mathcal{R}})$. 

Finally, $\bar{O}^{\mathcal{R}}$ is defined as the collection of the selected objects from each iteration $\bar{O}_{i}^{\mathcal{R}}$. As a result, we get that the set of selected objects considering steps 1 and 2 corresponds to $\bar{O} = \bar{O}^{\mathcal{T}} \cup \bar{O}^{\mathcal{R}}$. The complete process of obtaining $\bar{O}^{\mathcal{R}}$ is detailed in Algorithm~\ref{algorithm:LLM^R}.
\begin{algorithm}[h]
\caption{$LLM^{\mathcal{R}}$: Relationship-based selection}\label{algorithm:LLM^R}
\begin{algorithmic}[1]
\setstretch{1.07} 
\State \textbf{Require}: graph state $S$, selected objects $\bar{O}^{\mathcal{T}}$, goal $\mathcal{G}$, max. iterations $I$, and functions $\Phi$, $LLM^{\mathcal{R}}_{i} $
\State \textbf{initialize} $\bar{O}_{i}^{\scriptscriptstyle \mathcal{T} \cup \mathcal{R}} \leftarrow \bar{O}^{\mathcal{T}}$
\For{$i \in \{0, 1, \ldots, I\}$}
    \State \textbf{get connected graph} $S_{i}^{\mathcal{E}} \leftarrow \Phi(S, \bar{O}_{i}^{\scriptscriptstyle \mathcal{T} \cup \mathcal{R}})$
    \State \textbf{select nodes} $\bar{O}^{\mathcal{R}}_{i} \leftarrow LLM_{i}^{\mathcal{R}}(\bar{O}_{i}^{\scriptscriptstyle \mathcal{T} \cup \mathcal{R}}, S_{i}^{\mathcal{E}}, \mathcal{G})$
    \If{$\bar{O}^{\mathcal{R}}_{i}$ is not empty}
    \State \textbf{assign} $\bar{O}_{i}^{\scriptscriptstyle \mathcal{T} \cup \mathcal{R}} \leftarrow \bar{O}_{i}^{\scriptscriptstyle \mathcal{T} \cup \mathcal{R}}  \cup \bar{O}_{i}^{\mathcal{R}}$
    \Else
    \State \textbf{break}
    \EndIf
\EndFor
\State \textbf{output} $\bar{O}^{\mathcal{R}} \leftarrow \bar{O}_{i}^{\scriptscriptstyle \mathcal{T} \cup \mathcal{R}} \, \backslash \, \bar{O}^{\mathcal{T}}$
\end{algorithmic}
\end{algorithm}

\subsection{Integrating Taxonomy Graph States with LLMs}
So far, we have introduced functions based on LLMs that use information encoded in a graph to condition their output. Consequently, since LLMs expect natural language as input, the information stored in the graph must be presented in a natural language form.

For the function $LLM^{\mathcal{T}}$, it is only necessary to generate a string that lists the categories present at a given hierarchical level. For example, the following string could be included in the context window of the LLM: \emph{``Categories: electronics, lighting, appliances, \ldots"}. In the case of the lowest level, where object classes replace categories, the string could be: \emph{``Objects: apple, chair, toilet, \ldots"}.

For the relationship-based object step selection via $LLM^{\mathcal{R}}$, a similar structure is employed. Nevertheless, each component of the list consists of two objects and their corresponding relationship. For instance, \emph{``Relationships between objects: banana (34) is inside the fridge (70), plate (67) is on the table (45), \ldots"}. Here, the numbers in parentheses correspond to the unique IDs of individual objects. These are necessary because relationships pertain to specific objects rather than object classes. In this example, the distinction ensures that not every banana is assumed to be inside every fridge, but rather that only the banana with ID 34 is inside the fridge with ID 70.

\subsection{Planning with State Graphs}\label{sec:planning_w_graphs}
Graph-based state representations provide an intuitive, object-centric abstraction of an environment where actions represent operations on specific objects within this environment. More precisely, actions either change objects' attributes $h_{i}$ and/or modify relationships between objects $r_{i}$. For instance, when a robot takes the action ${a_{t} = \textit{close(microwave)}}$, it changes the microwave's attribute \emph{open} to \emph{closed}. Furthermore, the action $a_{t} = \textit{grab(apple)}$, when applied to an apple on a table, creates a new edge between the robot and the apple and removes the relationship the apple had with the table. This approach provides a compact way of representing transitions in the environment, where, by defining $S_{t}$ as the graph at a given time step, given the \emph{effects} an action has on its attributes and edges, we obtain the transition function ${S_{t+1}=f(S_{t}, a_{t})}$.

Moreover, attributes and edges are integral in constructing the affordance function $\bar{\mathcal{A}}(S_{t})$, as they are employed to define \emph{preconditions} required for an action to be applicable. Consider the action $a_{t} = \textit{close(microwave)}$: for this to be applicable, the microwave must have the attribute \emph{is open}, and the relationship \emph{near} must exist between the robot and the microwave, indicating that the microwave is within reach of the robot. 

As a result, by defining the actions in $\mathcal{A}$ along with their respective graph-based preconditions and effects, task planning problems can be addressed using graphs \cite{agia2022taskography,garrett2021integrated}. By employing the functions $f(S_{t}, a_{t})$ and $\bar{\mathcal{A}}(S_{t}, a_{t})$, given $S_{0}$ and $\mathcal{G}$, a policy can be derived by searching the state-action space.

In this work, we employ two types of policies. The first type follows traditional planning approaches, while the second is based on LLMs. These are described below.

\subsubsection{Search-based Policies}
These policies are derived from a solution to a search problem, given an initial condition $S_{0}$ and a set of goal states $\mathcal{G}$, e.g., Fast Downward \cite{helmert2006fast} or MCTS \cite{coulom2006efficient}. This solution is a plan $\tilde{a}=(a_{0}, \ldots, a_{n-1})$ leading to a goal state, that is, $f(a_{n-1}, S_{n-1})=S_{n} \in \mathcal{G}$. The plan is formulated by utilizing the known functions $f(S_{t}, a_{t})$ and $\bar{\mathcal{A}}(S_{t})$ to navigate the state-action space until a goal state $S_{n} \in \mathcal{G}$ is reached. In the context of robotics, the plan's initial condition can be estimated from the current state of the environment. However, the set $\mathcal{G}$ must be specified by some entity, e.g., a human. In practice, this is achieved by specifying desired Boolean conditions, for example, \emph{on(banana, kitchen table)=True and closed(microwave)=True}. As a result, $\mathcal{G}$ would include every state fulfilling these conditions.
%

\subsubsection{LLM-based Policies}\label{sec:llm-policies}
Multiple policies based on LLMs have been introduced \cite{wang2024survey}. In this work, we employ an approach consisting of three parts: a pre-trained LLM, context, and the affordance function $\bar{\mathcal{A}}(S_{t})$. The pre-trained model can be any existing LLM, such as GPT \cite{openai2023gpt4}. The context, provided in natural language, instructs the LLM to select the best action for a given state, aiming to move the agent towards a state that is closer to achieving the goal $\mathcal{G}$. To ground the actions selected by the LLM, at every time step, it is conditioned to choose actions solely from $\bar{\mathcal{A}}(S_{t})$. Consequently, for a given state $S_{t}$, the LLM-based policy outputs a feasible action to drive the environment toward a state closer to those in the set $\mathcal{G}$. This process is applied iteratively until the goal state is reached or the maximum planning length is attained.
\section{Experiments}\label{sec:experiments}
We empirically validated our method in three scenarios. Firstly, we studied the state-space reduction performance of our method. These experiments evaluate the main contribution of our work, which involves selecting only the necessary objects from an environment before executing planning algorithms. Nevertheless, the end goal of the work is to achieve better performance at the task planning level. Therefore, we also compared the performance of different planning methods, including both classical planning and LLM-based approaches, when the state space is reduced. Lastly, we validated the complete framework on a real-world robotic platform using a 7-DoF manipulator.
\begin{figure}[t]
    \centering
    \includegraphics[trim = {0 15 0 0}, clip,width=\columnwidth]{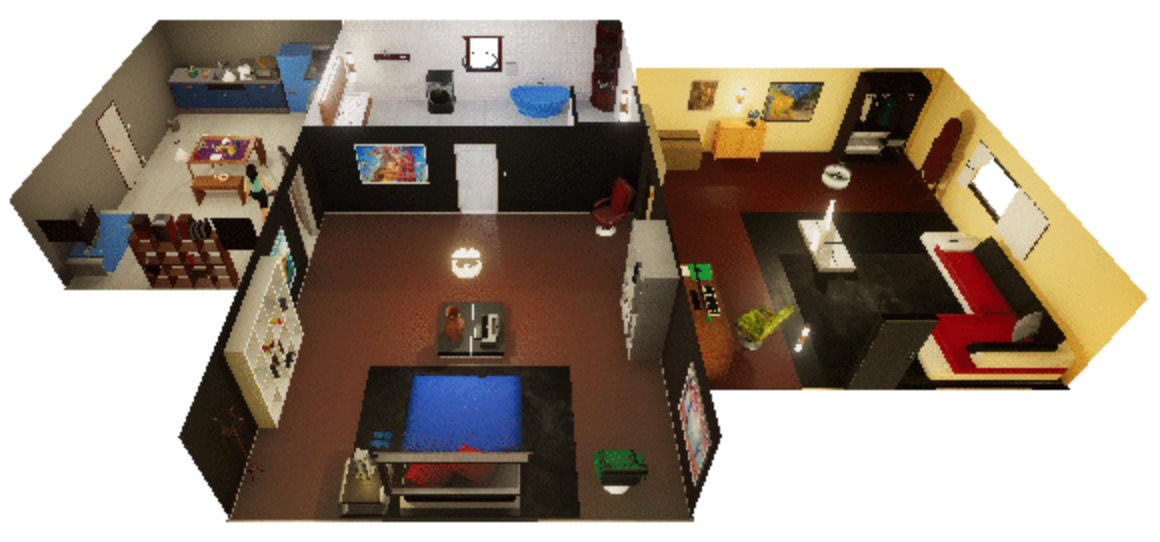}
    \caption{Example of a VirtualHome environment: The agent can navigate the house and interact with objects in multiple rooms. Six environments were employed, with the number of objects the agent could interact with ranging from 221 to 348. These objects have properties such as being \emph{openable}, \emph{grabbable}, or \emph{switchable}. Image extracted from \url{http://virtual-home.org/}.}
    \label{fig:virt_example}
\end{figure}

\begin{table*}[t]
\centering
\caption{Number of selected objects with GPT-4o and GPT-3.5 on successful trials. `N' denotes the number of subtasks.}
\begin{adjustbox}{max width=\textwidth}
\begin{tabular}{cccccccccccccc}
\toprule
N & \multicolumn{6}{c}{GPT-4o} & \multicolumn{6}{c}{GPT-3.5} \\
\cmidrule(lr){2-7} \cmidrule(lr){8-13}
 & \multicolumn{2}{c}{\# objects environment $|O|$} & \multicolumn{2}{c}{\# necessary objects $|O_\mathrm{g}|$} & \multicolumn{2}{c}{\# selected objects $|\bar{O}|$} & \multicolumn{2}{c}{\# objects environment $|O|$} & \multicolumn{2}{c}{\# necessary objects $|O_\mathrm{g}|$} & \multicolumn{2}{c}{\# selected objects $|\bar{O}|$} \\
\cmidrule(lr){2-3} \cmidrule(lr){4-5} \cmidrule(lr){6-7} \cmidrule(lr){8-9} \cmidrule(lr){10-11} \cmidrule(lr){12-13}
 & Mean & Std & Mean & Std & Mean & Std & Mean & Std & Mean & Std & Mean & Std \\
\midrule
1 & 278.64 & 45.56 & 4.21 & 3.66 & 8.50 & 8.03 & 273.17 & 44.93 & 4.50 & 4.21 & 6.67 & 5.37 \\
2 & 273.17 & 44.26 & 7.89 & 4.36 & 10.00 & 5.05 & 271.13 & 42.14 & 8.00 & 4.71 & 9.20 & 5.81 \\
3 & 276.89 & 46.57 & 10.26 & 4.31 & 12.53 & 4.34 & 270.50 & 50.45 & 9.43 & 3.88 & 12.00 & 4.98 \\
4 & 282.00 & 39.59 & 17.18 & 7.39 & 19.00 & 7.92 & 273.83 & 45.66 & 15.83 & 8.63 & 17.17 & 8.31 \\
5 & 275.79 & 46.71 & 19.83 & 7.73 & 21.71 & 8.06 & 265.85 & 40.16 & 18.39 & 8.90 & 19.77 & 9.10 \\
\bottomrule
\end{tabular}
\label{tab:size_reduction}
\end{adjustbox}
\end{table*}
\subsection{State-space size reduction}
To study the state-space size reduction performance of our method, we used \emph{VirtualHome} \cite{puig2018virtualhome} (see Fig.~\ref{fig:virt_example}). VirtualHome is a simulator that allows controlling agents in a household environment by sending them high-level commands, such as \emph{walk to kitchen} or \emph{open the fridge}. The agent is equipped with lower-level action primitives that enable it to execute these high-level commands. Consequently, it is an ideal setting for studying task-planning problems. We evaluated our method in six different household environments with $\sim 280$ objects each. Furthermore, to analyze the robustness of our method, we evaluated it for planning objectives that were increasingly difficult to achieve. To increase difficulty, we composed multiple single-objective tasks into single multi-objective tasks. For example, one single-objective task can be \emph{put a beer on a table}, and a multi-objective task composed of two subtasks would be \emph{put one beer on the kitchen table and put one chicken inside the microwave}.

\subsubsection{Unsuccessful state-space reductions}
\begin{figure}[t]
    \centering
    \includegraphics[trim = {10 20 10 17}, clip, width=\columnwidth]{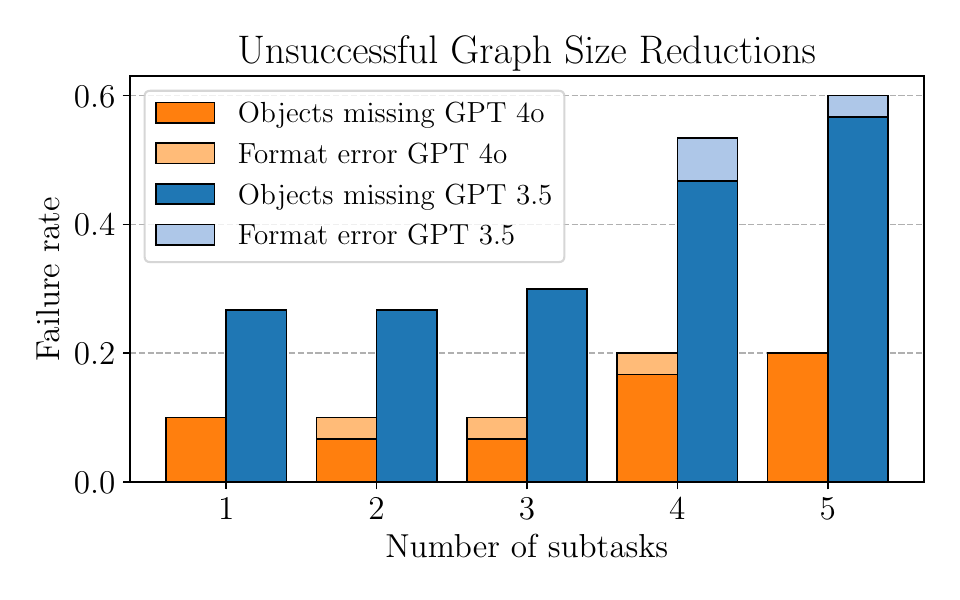}
    \caption{Performance comparison of state space size reduction using $\text{GPT-3.5}$ and GPT-4o for tasks containing one to five subtasks. \textit{Objects missing} indicates that some necessary objects for the task are not selected, while \textit{format error} means the LLM's output does not adhere to the required format.}
    \label{fig:size_recution_performance}
\end{figure}
We employed GPT-3.5 and GPT-4o as our pretrained LLMs. Fig.~\ref{fig:size_recution_performance} showcases their performance in successfully selecting objects for tasks containing 1 to 5 subtasks, with each case evaluated for 30 different objectives. In this figure, we analyze whether the selected objects contain all the necessary elements for creating a successful plan, i.e., $O_{\mathrm{g}} \subseteq \bar{O}$. If this condition is met after selecting the objects, the process is labeled as successful. Note that a selection process can also be unsuccessful if the LLM generates outputs that are not possible to process due to format errors. In such cases, the LLM is queried again. This re-query process occurs a maximum of three times, and if a format error is detected again after the third attempt, the object selection process is labeled as unsuccessful. For each task, the subset of necessary objects is defined based on a heuristic that exploits VirtualHome's structure. 

We can observe that as the number of subtasks per task increases, the performance of the object selection process decreases for both GPT-3.5 and GPT-4o. This result is not very surprising, as more subtasks signify a more complex object selection process. However, a more interesting observation is that GPT-4o significantly outperforms GPT-3.5, with failure rates ranging from 0.1 to 0.2, in contrast to 0.23 to 0.6. Since GPT-4o is a more powerful model than GPT-3.5, this suggests that as these models improve, they will become better at the object selection problem. This is without detracting from the current high performance of GPT-4o. Lastly, we can also observe that although failures did occur due to format errors, they were not very significant in the overall performance of the models.

\subsubsection{Size of successful reductions}
Apart from evaluating whether the selected objects contain $O_{\mathrm{g}}$, it is also important to study the size of the set of selected objects $\bar{O}$. To this end, Table~\ref{tab:size_reduction} provides the average number of selected objects for the \textbf{successful cases} of Fig.~\ref{fig:size_recution_performance}. This table also shows the average initial number of objects and the average number of necessary objects. For both GPT-4o and GPT-3.5, we can observe that $|\bar{O}|$ is similar to $|O_{\mathrm{g}}|$, especially when compared to the initial number of objects, which is orders of magnitude larger. It is interesting to note that in multiple cases, GPT-3.5 selects fewer objects than GPT-4o. At first glance, this might indicate better performance for GPT-3.5. However, this is because GPT-3.5 fails to properly extract some information from the task goal and graph during the object selection process, leading to more simplistic solutions that overlook relevant objects and, consequently, to the high failure rates observed in Fig.~\ref{fig:size_recution_performance}.

\begin{table}[t]
    \centering
    \footnotesize
    \caption{Comparison of success rates and average trajectory steps across different methods using the obtained reduced states.}
    \begin{tabular}{ccccccc}
        \toprule
        N & \multicolumn{2}{c}{GPT-4o} & \multicolumn{2}{c}{GPT-3.5} & \multicolumn{2}{c}{MCTS} \\
        & Succ. rate & Steps & Succ. rate & Steps & Succ. rate & Steps \\
        \midrule
        1 & \textbf{1.000} & 5.267 & 0.100 & 4.000 & \textbf{1.000} & 22.200  \\
        2 & \textbf{1.000} & 9.700 & 0.000 & \textbackslash{} & 0.700 & 46.571  \\
        3 & \textbf{0.850} & 13.933 & 0.000 & \textbackslash{} & 0.450 & 79.440 \\
        4 & \textbf{0.750} & 16.267 & 0.000 & \textbackslash{} & 0.400 & 99.500 \\
        5 & \textbf{0.733} & 22.765 & 0.000 & \textbackslash{} & 0.000 & \textbackslash{} \\
        \bottomrule
    \end{tabular}
    \label{tab:comparison}
\end{table}
\begin{figure*}[t]
    \centering
    \includegraphics[trim = {0 20 0 0}, clip, width=\textwidth]{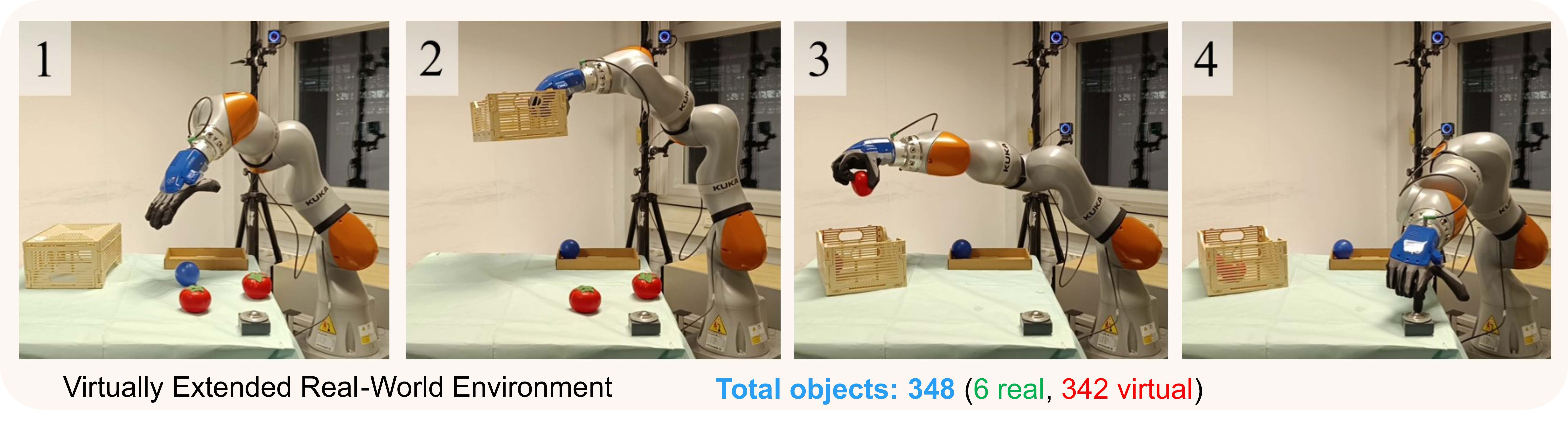}
    \caption{Example of a real-world execution of the proposed method. Objective: put the tomatoes inside the crate, the rotten tomatoes (blue ball) in the bin (box in the corner of the table) and press the button once you finish.}
    \label{fig:robot_exps}
\end{figure*}
\subsection{Planning}
Given that the objective of reducing the state space size is to simplify the planning problem, in Table~\ref{tab:comparison}, we provide the performance of LLM-based planners and a search-based approach, namely, Monte Carlo Tree Search (MCTS) \cite{coulom2006efficient}, in reduced-state-space settings. The employed LLM-based planners operate on a per-time-step basis, as explained in Section~\ref{sec:planning_w_graphs}. \textbf{Every planner was executed over the reduced states obtained using GPT-4o}.

\subsubsection{Planning performance with state space reduction} 
From Table~\ref{tab:comparison}, we observe that GPT-4o outperforms $\text{GPT-3.5}$ and MCTS in planning, achieving a 0.73 success rate even in the most challenging scenarios. The performance of GPT-3.5 is surprisingly low; we noted that this occurred because the model could not properly reason from the provided context, thereby selecting actions that did not bring the agent closer to the goal. Lastly, MCTS obtained the expected performance: as the problem size increases, it becomes increasingly intractable. It is also worth noting that the number of steps in the solutions found by MCTS is considerably larger than those found by the LLM-based planners. This result is expected, as the primary objective of MCTS is to find a solution, not necessarily \emph{the best} solution, which can lead to redundancies or unnecessary steps. In contrast, LLM-based planners provided solutions guided by their commonsense knowledge.

\subsubsection{Planning performance without state space reduction} 
The same experiments shown in Table~\ref{tab:comparison} were conducted for MCTS without reducing the state space size. As expected, \textbf{MCTS was unable to find any solutions} in these scenarios, not even for problems with only one subtask, highlighting the importance of reducing the state space size. For LLM-based planners, obtaining such results was not feasible, as planning with such long context windows would be prohibitively expensive. This further demonstrates the advantage of reducing the state space size.

\subsubsection{Overall performance}
\begin{figure}[t]
    \centering
    \includegraphics[trim = {10 20 10 17}, clip,width=\columnwidth]{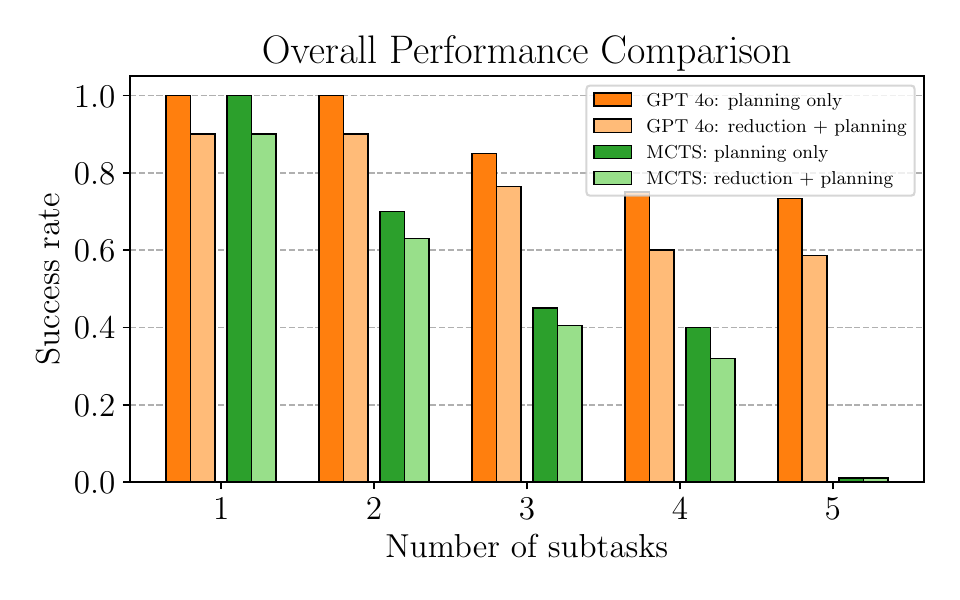}
    \caption{Evaluating the planning performance of two scenarios: 1) the complete framework, incorporating state reduction and planning, and 2) assuming zero failures in state reductions.}
    \label{fig:planning_success_rate}
\end{figure}

Finally, Fig.~\ref{fig:planning_success_rate} illustrates the overall performance of the top-performing planning agents: GPT-4o and MCTS. The overall performance encompasses the reduction of state space size and planning processes. Consequently, if the state space size reduction process fails, the planner fails automatically, as it would not be possible to find a solution. This figure also compares the planning-only performance with the overall performance. From this, we can conclude that although the overall performance is naturally lower than the planning-only performance, the difference is not drastic, indicating that the proposed framework offers a viable approach for addressing large-scale planning problems. Once again, we observe that $\text{GPT-4o}$ outperforms MCTS, and the performance decreases as the number of subtasks increases. 

\subsection{Real-world validation}
We conducted a real-world validation of the proposed method, utilizing the best-performing models from the previous subsections, namely GPT-4o, for both state space selection and planning. The objective of this experiment is to demonstrate the applicability of the proposed approach to real-world problems and its potential, given an extensive library of motion primitives. Fig.~\ref{fig:robot_exps} presents a sequence of images depicting the task being executed on the robotic setup. It should be noted that, for practical reasons, the real setup did not include hundreds of objects; instead, the state space was virtually extended to also incorporate all objects from one of the VirtualHome simulated environments. As a result, the state space reduction process included $\sim350$ objects.

The following subtasks were used to establish planning goals: 1) \emph{put a tomato inside the crate}, 2) \emph{put a rotten tomato inside the bin}, and 3) \emph{press the button}. Each of these subtasks was tested individually and then incrementally combined until achieving the most complex task in this setup: \emph{Put the tomatoes inside the crate and the rotten tomatoes inside the bin. Once you have completed these steps, press the button}. Note that this task requires consideration of temporal information, as it is only correct to press the button after completing the other subtasks. This makes the planning problem more challenging for a classical planning approach, as it increases the size of the search space. Nevertheless, the LLM-based planner showed no limitations when solving this problem. 

Motion primitives were designed to accomplish all the required high-level actions dictated by the task planner. The most notable motion primitive is the one that flips the crate, which is a necessary step because the crate is initially upside down. This primitive is significantly more challenging to achieve than simpler primitives, such as pick and place, as it requires complete pose control while interacting with other objects, such as the crate and table. For more details on the real-world experiments, the reader is referred to the attached video: \url{https://youtu.be/6ro2UOtOQS4}.
\section{Conclusions}\label{sec:conclusions}
In this work, we present a method to reduce the state space in large-scale task planning by combining LLMs with state graphs and object taxonomies. We extensively evaluated the method using the VirtualHome simulator, performing tasks that would have been unfeasible with both search-based and LLM-based planners. With a reduced state space, LLM planners handled planning problems more cost-effectively, and GPT-4o significantly outperformed GPT-3.5, suggesting further improvements with model evolution. MCTS, a classical planner, also benefited by being able to solve an increased number of problems after reducing the state space size.
We also validated our method on a real-world robotic platform, proving its practical use.

\section*{\large Funding}
This project is made possible by a contribution from the National Growth Fund program NXTGEN Hightech.


\medskip

%
\bibliographystyle{IEEEtran}  
\bibliography{IEEEexample}  








\end{document}